\documentclass[sigconf]{acmart}

\usepackage{booktabs}
\usepackage{multirow}
\usepackage{graphicx}	
\usepackage{color, colortbl}
\usepackage{subcaption}
\usepackage{enumitem}
\usepackage{amsthm}
\usepackage{listings}

\usepackage{pgfplots}
\usetikzlibrary{patterns}
\usetikzlibrary{plotmarks}
\pgfplotsset{compat=1.11,
    /pgfplots/ybar legend/.style={
        /pgfplots/legend image code/.code={%
            \draw[##1,/tikz/.cd,bar width=3pt,yshift=-0.2em,bar shift=0pt]
            plot coordinates {(0cm,0.8em)};
        },
    },
}

\theoremstyle{definition}
\newtheorem{definition}{Definition}

\usepackage{xcolor}

\definecolor{codegreen}{rgb}{0,0.6,0}
\definecolor{codegray}{rgb}{0.5,0.5,0.5}
\definecolor{codepurple}{rgb}{0.58,0,0.82}
\definecolor{backcolour}{rgb}{0.95,0.95,0.92}

\definecolor{wfblue}{RGB}{180, 199, 231}
\definecolor{wfgreen}{rgb}{0.858, 0.188, 0.478}
\definecolor{wfred}{rgb}{0.858, 0.188, 0.478}
\definecolor{wfgray}{rgb}{0.858, 0.188, 0.478}

\lstdefinestyle{mystyle}{
    commentstyle=\color{codegreen},
    keywordstyle=\color{magenta},
    numberstyle=\footnotesize\color{codegray},
    stringstyle=\color{codepurple},
    basicstyle=\ttfamily\footnotesize,
    breakatwhitespace=false,         
    breaklines=true,                 
    captionpos=b,                    
    keepspaces=true,                 
    numbers=left,                    
    numbersep=5pt,                  
    showspaces=false,                
    showstringspaces=false,
    showtabs=false,                  
    tabsize=2
}

\newcommand{\ourmethod}{\textsc{Mango}}
\newcommand{\ourmethodtop}{\ourmethod$^{\textit{top}}$}
\newcommand{\ourmethodfull}{\ourmethod$^{\textit{full}}$}

\newcommand{\ourgpt}{GPT + CCSK}

\newcommand{\website}{\url{https://mango.mpi-inf.mpg.de}}

\newcolumntype{G}{>{\columncolor [gray] {0.90}}c}

\newcommand{\squishlist}{
   \begin{list}{•}
    { 
        \setlength{\itemsep}{0pt}
        \setlength{\parsep}{0pt}
        \setlength{\topsep}{0pt}       \setlength{\partopsep}{0pt}
        \setlength{\leftmargin}{1.5em} \setlength{\labelwidth}{1em}
        \setlength{\labelsep}{0.5em} 
    } 
}

\newcounter{Lcount}
\newcommand{\squishlisttwo}{
    \begin{list}{\arabic{Lcount}. }
    { 
        \usecounter{Lcount}
        \setlength{\itemsep}{0pt}
        \setlength{\parsep}{0pt}
        \setlength{\topsep}{0pt}       \setlength{\partopsep}{0pt}
        \setlength{\leftmargin}{2em} \setlength{\labelwidth}{1em}
        \setlength{\labelsep}{0.5em} 
    }
}

\newcommand{\squishend}{
    \end{list}
}

\copyrightyear{2024}
\acmYear{2024}
\setcopyright{rightsretained}
\acmConference[CIKM '24]{Proceedings of the 33rd ACM International Conference on Information and Knowledge Management}{October 21--25, 2024}{Boise, ID, USA}
\acmBooktitle{Proceedings of the 33rd ACM International Conference on Information and Knowledge Management (CIKM '24), October 21--25, 2024, Boise, ID, USA}
\acmDOI{10.1145/3627673.3679768}
\acmISBN{979-8-4007-0436-9/24/10}

\makeatletter
\gdef\@copyrightpermission{
  \begin{minipage}{0.3\columnwidth}
   \href{https://creativecommons.org/licenses/by/4.0/}{\includegraphics[width=0.90\textwidth]{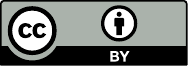}}
  \end{minipage}\hfill
  \begin{minipage}{0.7\columnwidth}
   \href{https://creativecommons.org/licenses/by/4.0/}{This work is licensed under a Creative Commons Attribution International 4.0 License.}
  \end{minipage}
  \vspace{5pt}
}
\makeatother

\begin{document}

\title{Cultural Commonsense Knowledge for Intercultural Dialogues}


\author{Tuan-Phong Nguyen}
\orcid{0000-0002-1674-2741}
\affiliation{%
  \institution{Max Planck Institute for Informatics}
  \city{Saarbr\"ucken}
  \country{Germany}}
\email{tuanphong@mpi-inf.mpg.de}

\author{Simon Razniewski}
\orcid{0000-0002-5410-218X}
\affiliation{%
  \institution{TU Dresden and ScaDS.AI}
  \city{Dresden/Leipzig}
  \country{Germany}}
\email{simon.razniewski@tu-dresden.de}

\author{Gerhard Weikum}
\orcid{0000-0003-4959-6098}
\affiliation{%
  \institution{Max Planck Institute for Informatics}
  \city{Saarbr\"ucken}
  \country{Germany}}
\email{weikum@mpi-inf.mpg.de}


\begin{abstract}
Despite recent progress, large language models (LLMs) still face the challenge of appropriately reacting to the intricacies of social and cultural conventions. 
This paper presents \ourmethod{}, a methodology for distilling high-accuracy, high-recall assertions of cultural knowledge. We judiciously and iteratively prompt LLMs for this purpose from two entry points, concepts and cultures. Outputs are consolidated via clustering and generative summarization.
%
Running the \ourmethod{} method with GPT-3.5 as underlying LLM yields 167K high-accuracy assertions for 30K concepts and 11K cultures, 
surpassing prior resources by a large margin in quality and size.
%
In an extrinsic evaluation for intercultural dialogues,
we explore augmenting dialogue systems with cultural knowledge assertions. 
Notably, despite LLMs inherently possessing cultural knowledge, 
we find that adding knowledge from \ourmethod{} improves the overall quality, specificity, and cultural sensitivity of dialogue responses, as judged by human annotators.
Data and code are available for download\footnote{\website{}}.
\end{abstract}




\maketitle

\section{Introduction}

\paragraph{Motivation}
The theme of enhancing computers with commonsense knowledge (CSK) has been revived in recent years, to improve the robustness and versatility of AI-powered NLP methods (e.g., \cite{bosselut2019comet, comet-atomic-2020, west2021symbolic, 10.1162/daed_a_01906}). CSK collections, such as ConceptNet~\cite{speer2017conceptnet} and ATOMIC~\cite{atomic}, capture assertions about mundane objects and human activities in a broad sense, so as to provide universal background knowledge.
However, norms for human behavior are often diverse, with substantial diversity across different parts of the world and different socio-cultural groups.
For example, when two people meet, they shake hands in the Western world, but this would be uncommon in 
Southeast Asia (where people \textit{wai} each other), and teenagers have yet another way of greeting.
Therefore, to adapt AI applications to specific user contexts, the goal of this research is to capture {\em culturally aware} commonsense knowledge, {\em CCSK} for short. 

An NLP task where cultural knowledge can play a game-changing role is human-computer dialogue, where the system should consider the human's cultural background. An even more challenging case is conversations between two partners from different cultures; we call this case {\em intercultural dialogue}.
Computers can take one, both or none of the roles in this setting; in the latter case, it could still be important to have cultural knowledge to analyze and post-process the dialogue (e.g., generate a de-briefing summary).


\paragraph{State of the Art and its Limitations}
On this nascent direction, resources for cultural knowledge have limited scope and scale (e.g., \citet{acharya2020towards}, StereoKG~\cite{stereokg}, Candle~\cite{candle}). The by far largest of these is the Candle knowledge base, 
which comprises ca. 60K assertions automatically distilled from nearly 1M sentences mined from a large web crawl. 
This is still insufficient in coverage of concepts and cultural groups.
Note that cultures are not just geo-regions, but should ideally consider also demographic and social traits of the respective groups (e.g., female teenagers in Korea, French people of North-African descent, etc.).

An obvious idea is to abandon explicit assertions, and instead rely on the latent knowledge provided by large language models (LLMs). 
Indeed, LLMs have shown promising performance on tasks that require understanding the personal background of users in conversational settings \cite{chen2024large}.
However, there are severe disadvantages:
(i) strong dependence on the quality of the training corpus (which may 
exhibit certain biases and risks of hallucination); 
(ii) lack of transparency of what piece of knowledge is utilized in which context;
(iii) as a result, lack of controllability for programmed applications; 
and
(iv) lack of interpretability by end-users.
For these reasons, our approach is to construct transparent and tangible collections of explicit assertions, which can later be judiciously injected into LLMs for specific tasks.

\paragraph{Challenges and Research Questions}
The twofold challenge is to expand the {\em coverage} of cultural groups and culture-specific assertions, while maintaining or even improving the {\em quality} of the assertions.
This poses three research questions:
\begin{itemize}
    \item \textbf{RQ1:} How can we systematically capture more and diverse cultural groups?
    \item \textbf{RQ2:} How can we substantially enlarge the amount of captured assertions, going beyond single-sentence retrieval from the web?
    \item \textbf{RQ3:} How can we ensure the specificity of assertions, while avoiding undue stereotypes and limiting redundancy due to frequency bias?
\end{itemize}

\paragraph{Approach and Contribution}
The methodology devised in this work, called \ourmethod{}
(\underline{m}ethod for cultur\underline{a}l commo\underline{n}sense knowled\underline{g}e distillati\underline{o}n),
comprises two stages: a coverage-oriented  stage for {\em assertion generation}, and a quality-oriented stage for {\em assertion consolidation}. The first stage addresses RQ1 and RQ2, and the second addresses RQ3.

To enlarge the pool of candidates, we leverage 
a large language model (LLM), specifically GPT-3.5, to generate assertions. Unlike prior works that tapped into web crawls, this is an implicit way of tapping the LLM's pre-training collection. Compared to web retrieval, the scales are similar but the advantage is that LLM training data involves efforts to remove spam and offensive content.
Asking an LLM to generate assertions requires judicious prompting. For high coverage of diverse cultures, an important novelty in our method is to construct prompts for both concepts (incl. human activities) from an existing (culturally agnostic) commonsense knowledge base  and groups from a large pool of cultures.

To ensure the informativeness of assertions and to tame the redundancy of generated candidates (RQ3), the second stage consists of steps for assertion consolidation.
This is carried out by clustering the pool of assertions
into topically and culturally coherent groups, considering both the key concept in an assertion and the culture to which it refers. The LLM is leveraged again to generate a concise summary statement for each cluster. This exploits the LLM's language skills, but does not rely on actual knowledge by the LLM.

The main contributions of this work are:

\begin{enumerate}
    \item The \ourmethod{} methodology for efficiently distilling CCSK from LLMs, at high precision and recall.
    \item Running the \ourmethod{} method with GPT-3.5, to obtain 167K assertions for 30K concepts and 11K cultures, substantially surpassing prior CCSK resources in size and quality.
    \item An extrinsic evaluation for intercultural dialogues, where the injection of  \ourmethod{} assertions significantly improves the specificity and cultural sensitivity of responses.
\end{enumerate}

We have made our data and code publicly available at \website{}.
    
\section{Related Work}

\subsection{Commonsense Knowledge Acquisition}
Commonsense knowledge (CSK)  has a long tradition in AI~\cite{lenat1995cyc, liu2004conceptnet, DBLP:conf/aaai/GordonDS10}.
Early
projects such as
Cyc~\cite{lenat1995cyc} and ConceptNet~\cite{liu2004conceptnet, speer2017conceptnet}
rely
on human contributors and 
focus
on collecting CSK about everyday concepts.
This approach has been revived in the ATOMIC project~\cite{atomic}, which collects CSK about social interactions and events.

Another line of CSK acquisition methods is based on text extraction, for example,
WebChild~\cite{webchild}, 
Quasimodo~\cite{quasimodo}, 
TransOMCS~\cite{transomcs}, and
Ascent~\cite{ascent,ascentpp}.
These are more scalable, but contain noise and have lower assertion quality.

With the advent of LLMs, methods have been devised for generating CSK assertions from language models 
(e.g., \cite{bosselut2019comet,petroni2019language,comet-atomic-2020,west2021symbolic,cohen-etal-2023-crawling,west-etal-2023-novacomet}).
None of these captures culture-aware knowledge, though.

\subsection{Cultural Knowledge Acquisition}
There are only few works on culture-aware knowledge:
\citet{Anacleto2006}, \citet{acharya2020towards}, \citet{shwartz-2022-good}, StereoKG~\cite{stereokg}, Candle~\cite{candle}, NormsKB~\cite{fung-etal-2023-normsage}, and GD-COMET~\cite{bhatia-shwartz-2023-gd}.
These methods
are based on extraction from text~\cite{stereokg,shwartz-2022-good,candle}, norm discovery from multi-lingual conversations~\cite{fung-etal-2023-normsage}, or fine-tuning LLMs on cultural knowledge bases~\cite{bhatia-shwartz-2023-gd}.
The resulting assertions either suffer from high noise or limited coverage, with the largest resource, Candle, providing 60K assertions.
Most recently, \citet{ch-wang-etal-2023-sociocultural} aligned social situations from an English knowledge base~\cite{forbes-etal-2020-social} and Chinese QA forums to mine 3,069 social norms for these two cultures.
All of these prior works are very limited in scope and scale.


\subsection{Culture-Aware NLP}
Several benchmarks exist for tasks that involve cultural knowledge beyond the Western world, e.g.,
MaRVL~\cite{liu-etal-2021-visually}, 
GD-VCR~\cite{yin-etal-2021-broaden},
GeoMLAMA~\cite{geomlama}, 
CALI~\cite{huang-yang-2023-culturally},
and
FORK~\cite{palta-rudinger-2023-fork}.
Some methods (e.g., \cite{li-zhang-2023-cultural}, \cite{zhou-etal-2023-cross}) have used LLMs to solve tasks of this kind.
Most recently, contemporaneous works construct datasets to benchmark cultural knowledge and reasoning skills of LLMs, via extracting cultural norms and norm violations from Wikipedia texts~\cite{fung2024massively}, and AI-assisted interactive red-teaming~\cite{chiu2024culturalteaming}.

Harnessing explicit CSK assertions for dialogue systems has been studied in \cite{kim-etal-2023-soda}, based on the ATOMIC resource (about social interactions).
NormDial~\cite{li-etal-2023-normdial} is a small collection of 267 human-written social norms specific to America and China, used to generate a bilingual dialogue dataset on social norms adherence or violation.
In this work, we 
aim at a much broader coverage of diverse cultures, and apply our large-scale \ourmethod{} collection to the task of intercultural dialogues.


\begin{figure*}[t]
    \centering
    \includegraphics[width=\textwidth]{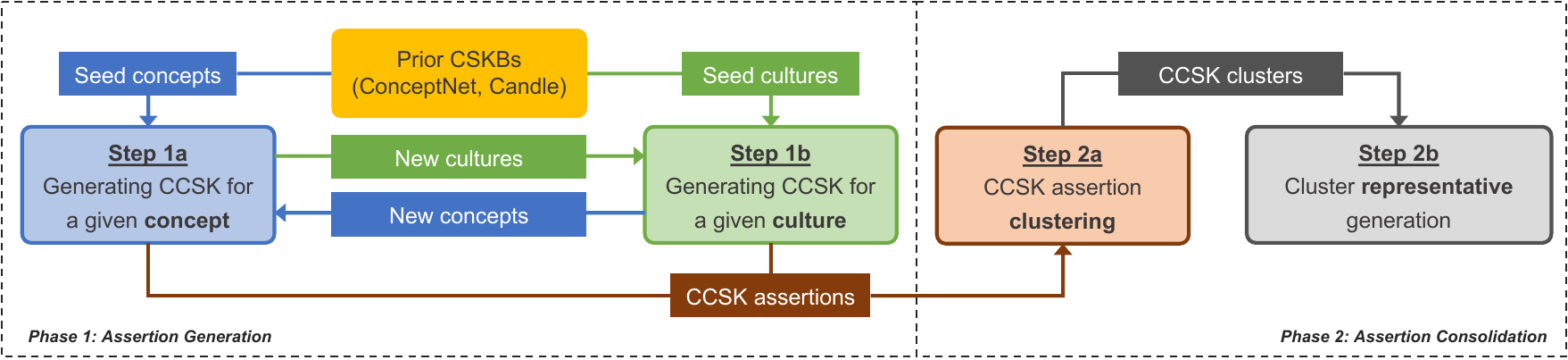}
    \caption{The \ourmethod{}'s CCSK distillation workflow: In Phase 1 (Assertion Generation), we use GPT-3.5 to generate CCSK assertions from two entry points, concepts and cultures, separately. In Phase 2 (Assertion Consolidation), we group semantically similar assertions into clusters, then generate a representative statement for each cluster. We provide examples in Table~\ref{tab:examples}.}
    \Description{The \ourmethod{} workflow consists of two phases: Assertion Generation and Assertion Consolidation.}
    \label{fig:method-overview}
\end{figure*}

\begin{table}[t]
    \centering
    \footnotesize
    \begin{tabular}{p{.93\columnwidth}}
        \toprule
        {\cellcolor[gray]{.9}
        \textbf{Step 1a: Generating contrastive CCSK for a given concept}} \\
        \textbf{Input concept:} chopsticks \\
        \textbf{Prompt:} You are a helpful assistant that writes culture-specific commonsense assertions. Some examples assertions are listed below: \\
        * car | Important in US, Germany | Considered luxury item in poorer countries \\
        * \texttt{<4 more examples...>} \\
        Please write assertions for the concept: chopsticks. \\
        \textbf{Parsed output:} \\
        • Concept: \textit{chopsticks}. Culture: \textit{Japan}. Statement: \textit{Standard eating utensils.} \\
        • Concept: \textit{chopsticks}. Culture: \textit{Western countries}. Statement: \textit{Considered exotic and less commonly used for everyday meals}. \\
        \midrule
        {\cellcolor[gray]{.9}\textbf{Step 1b: Generating contrastive CCSK involving a given culture}} \\
        \textbf{Input culture:} Japan \\
        \textbf{Prompt:} You are a helpful assistant that writes culture-specific commonsense assertions. Some examples assertions are listed below: \\
        * rice | Staple food in East Asia | Side dish in European countries \\
        * \texttt{<4 more examples...>} \\
        Please write assertions where one of the cultures is: Japan. \\
        \textbf{Parsed output:} \\
        • Concept: \textit{tipping}. Culture: \textit{Japan}. Statement: \textit{Not a common practice}. \\
        • Concept: \textit{tipping}. Culture: \textit{USA}. Statement: \textit{Common and expected practice in the service industry.} \\
        \midrule
        {\cellcolor[gray]{.9}\textbf{Step 2a: CCSK assertion clustering}} \\
        \textbf{Input:} >500K CCSK assertions \\
        \textbf{Sample output cluster:} \\
        • Concept: \textit{tipping}. Culture: \textit{Japanese}. Statement: \textit{Not a common practice}. (Frequency: 5) \\
        • Concept: \textit{leaving tip}. Culture: \textit{Japanese culture}. Statement: \textit{Not a common practice and may even be seen as rude.} (Frequency: 2) \\
        • Concept: \textit{tipping at restaurants.} Culture: \textit{Japan}. Statement: \textit{Tipping is not commonly practiced and can even be considered rude as it implies that the service is not already included in the price.} (Frequency: 1) \\
        • Concept: \textit{tipping service staff}. Culture: \textit{Japan}. Statement: \textit{Not a common practice and can even be considered rude or disrespectful.} (Frequency: 1) \\
        \midrule
        {\cellcolor[gray]{.9}\textbf{Step 2b: Cluster representative generation}} \\
        \textbf{Prompt:} Please generate a representative sentence for the following assertions: \\
        \texttt{<the sample cluster above>} \\
        \textbf{Output:} Concept: \textit{tipping}. Culture: \textit{Japan}. Statement: \textit{Tipping is not a common practice in Japan and can be considered rude or impolite.} (Frequency: 9) \\
        \bottomrule
    \end{tabular}
    \caption{Examples of the \ourmethod{} workflow. 
    }
    \label{tab:examples}
\end{table}

\section{Cultural Commonsense Knowledge Distillation}
We propose \ourmethod{}, a workflow for distilling and consolidating cultural commonsense knowledge (CCSK) from large language models (LLMs).

\begin{definition}[Cultural commonsense knowledge assertion]
    A cultural commonsense knowledge (CCSK) assertion $a = (c; g; s)$ is a triple of a concept $c$; a culture/cultural group $g$; and a natural-language statement $s$ that represents a cultural belief, norm, or common practice about the concept $c$ that applies to group $g$.
    An example CCSK assertion is: (\textit{tipping}; \textit{Japan}; \textit{Not a common practice}).
\end{definition}

The goal of \ourmethod{} is to generate CCSK for diverse cultures, covering a wide variety of concepts. Moreover, by clustering similar assertions, we obtain ranking signals for the assertions, which can be useful for downstream applications.

Our workflow consists of two phases, each consisting of two steps: (1a) generating CCSK for a given concept, (1b) generating CCSK for a given culture (Section~\ref{subsec:method-assertion-generate}), (2a) clustering similar CCSK assertions, and (2b) generating cluster representatives (Section~\ref{subsec:method-assertion-consolidate}). 
An overview of our workflow is depicted in Figure~\ref{fig:method-overview}.
We also provide input and output examples for each step in Table~\ref{tab:examples}.

\subsection{Phase 1: Assertion Generation}
\label{subsec:method-assertion-generate}

People of different cultures may have different perspectives on certain concepts.
These differences are what we are interested in, as they are crucial in situations where cross-cultural knowledge is required to understand one another.
One might ask LLMs to generate CCSK for a given pair of concept \emph{and} culture (e.g., prompting for perceptions of \textit{tipping} in \textit{Japan}).
However, the main drawback of this approach is that it requires concept and culture pairings, which is problematic because of the large number of possible combinations, and the fact that some combinations may not make sense.

Instead, we propose to use LLMs to generate CCSK for a given concept or a given cultural group \emph{separately}. 
This way we let the model decide which concepts are relevant for a given cultural group and vice versa, which will reduce the chance of nonsensical concept-culture combinations, hence reducing costs. 
Example prompts and outputs can be seen in Table~\ref{tab:examples}.

For each prompt, we provide the LLM with five pairs of example assertions randomly drawn from a set of human-written CCSK assertions.
Each example consists of a concept, and two different perspectives on that concept associated with two or more different cultures.
We implement our workflow using GPT-3.5.
However, in practice, other LLMs could be used.
We also run each prompt several times at a high temperature (i.e., high creativity), that can hopefully lead to different output assertions given the same input.

\paragraph{Step 1a: Generating CCSK for a given concept}
We seed this step with concepts from ConceptNet~\cite{speer2017conceptnet}, which is a popular commonsense knowledge base constructed by human annotations and consists of a wide range of everyday concepts.
In addition, we also use cultural concepts from Candle~\cite{candle} as inputs for this step.
Candle is a cultural knowledge base constructed by extracting relevant sentences from a filtered web crawl.

\paragraph{Step 1b: Generating CCSK for a given culture}
The seed cultures used for this step are taken from the Candle dataset, which include geo-locations (countries, continents, geo-regions), and religions.

\paragraph{Iterative Generation}
Step 1a and Step 1b are processed independently, and each step can generate new cultural groups and concepts, respectively.
These new concepts and cultural groups can be fed back to the corresponding approach in order to generate new CCSK assertions.

\subsection{Phase 2: Assertion Consolidation}
\label{subsec:method-assertion-consolidate}

It is natural that our CCSK generation approaches produce similar assertions that have very little or zero semantic differences.
The repetitions can come from the fact that there exist concepts with similar meanings (e.g., \textit{tea}, \textit{drinking tea}, and \textit{tea drinking}), and terms referring to the same groups (e.g., \textit{East Asia}, \textit{East Asian cultures}, and \textit{East Asian countries}) in the input.
Furthermore, as we run each prompt several times, even at high output randomness (temperature), repetitions are inevitable.

While untreated redundancy is undesirable, we propose to take advantage of these redundancies in order to rank our assertions.
Specifically, by grouping together semantically similar assertions, we create frequency signals, which can be useful for downstream applications which only look for a subset of highest-significance assertions.

\paragraph{Step 2a: Assertion Clustering}
As we generate hundreds of thousands of assertions in Phase 1, it is prohibitively expensive to run a clustering algorithm on all assertions at once.
It would even be infeasible for algorithms such as the Hierarchical Agglomerative Clustering (HAC) algorithm to process such a large amount of data points.
Instead, we propose a divide-and-conquer approach to clustering this large set of assertions.
Our approach consists of three substeps:
\begin{enumerate}
    \item \textbf{Clustering similar concepts:} The concepts in the assertions generated in Phase 1 are clustered into groups of similar concepts.
    \item \textbf{Clustering similar cultures:} Similarly to concept clustering, we group together similar cultures.
    \item \textbf{Clustering subsets of assertions:} For each pair of concept cluster and culture cluster, we only cluster the subset of assertions associated with any of the concepts and any of the cultures in those clusters.
\end{enumerate}

As the sizes of the entire concept and culture sets and the corresponding assertion subsets are substantially smaller than that of the entire set of generated assertions, standard clustering algorithms can process them efficiently.
In our experiment, we use SentenceBert embeddings~\cite{sbert} and the HAC algorithm for all 3 substeps.
Nevertheless, in practice, other text embedding models and clustering algorithms could be used.
An example assertion cluster can be found in Table~\ref{tab:examples}, in which the respective concept cluster includes \textit{tipping}, \textit{leaving tip}, \textit{tipping at restaurants}, \textit{tipping service staff}, and the culture cluster includes \textit{Japan}, \textit{Japanese}, \textit{Japanese culture}.

\paragraph{Step 2b: Representative Generation}
For each cluster of CCSK assertions, we generate a \emph{full sentence} as the representative of the cluster.
We also obtain the frequency of each cluster by adding up all the frequencies of its member assertions.

For the final output, we store a set of CCSK cluster instances, each containing the following information:
\emph{concept} (the representative concept of the CCSK cluster),
\emph{culture} (the representative culture of the CCSK cluster),
\emph{statement} (the representative sentence of the CCSK cluster),
\emph{similar statements} (the set of the statements of the member assertions),
and
\emph{frequency} (the frequency sum of the cluster) - see examples in Table~\ref{tab:examples}.

\section{Implementation of \ourmethod{}}

\begin{table}
\small
\centering
    \begin{tabular}{lll}
    \toprule
    \textbf{Step} & \textbf{Input} & \textbf{Output} \\
    \midrule
    \textbf{\begin{tabular}[c]{@{}l@{}}Step 1a\\ (2 runs)\end{tabular}} & 31,196 concepts & 468,543 assertions \\
    \midrule
    \textbf{\begin{tabular}[c]{@{}l@{}}Step 1b\\ (2 runs)\end{tabular}} & 10,796 cultures & 115,220 assertions \\
    \midrule
    \textbf{Step 2a} & 507,780 filtered assertions & 167,396 assertion clusters\\
    \midrule
    \textbf{Step 2b} & 167,396 assertion clusters & 167,396 full CCSK sentences \\
    \bottomrule
    \end{tabular}
\caption{Input and output of each step in our implementation.}
\label{tab:input-output-size}
\end{table}

We implement our workflow using GPT-3.5\footnote{Model name: \textit{gpt-3.5-turbo-1106}
}.
In the generation phase, we run each prompt five times with a temperature of 1.0 (high creativity).
For convenience, we ask the model to structure its responses into JSON objects\footnote{\url{https://platform.openai.com/docs/api-reference/chat/create\#chat-create-response_format}}.
In each prompt, we provide five CCSK assertion pairs as in-context learning examples, which are randomly sampled from a pool of ten examples written by humans.
We summarize the input and output of our implementation in Table~\ref{tab:input-output-size}.

\subsection{Input Concepts and Cultures}
We run each step in Phase 1 twice, whereas the first run is fed with seed concepts and seed cultures from prior datasets, and the second run is fed with new concepts and cultures generated in the first run.
\begin{itemize}
    \item \textbf{Seed concepts:} We take 8,357 concepts from Candle~\cite{candle}, which have more than 1 assertion, and 16,480 concepts from ConceptNet~\cite{speer2017conceptnet}, which have more than 3 assertions. We clean this set of concepts by filtering out incomprehensible words and phrases using GPT-3.5, retaining a seed set of 19,940 concepts for the first run of Step 1a.
    \item \textbf{Seed cultures:} We take 286 groups (272 geo-locations and 14 religions) from Candle as seed cultures for the first run of Step 1b.
\end{itemize}
After the first run of Step 1a with the seed concepts, we obtain 10,510 new cultures.
After the first run of Step 1b with the 286 seed cultures, and the second run with the new 10,510 cultures, we obtain 11,256 new concepts, which are used for the second run of Step 1a.

\subsection{Assertion Filtering}
After Phase 1, we obtain 581,563 distinct CCSK assertions.
We perform various simple filters to clean these assertions:
\begin{itemize}
    \item Removing too long or too short assertions: We restrict the length of assertion to be between 2 and 25 words.
    \item Removing assertions that have more than one sentence in the generated statements.
    \item Removing assertions whose cultures are too general or noisy: We construct a dictionary of words that a valid culture should not contain. For example, we filter out cultures such as ``Other cultures'' (context-dependent), ``Non-European countries'', ``Some parts of Asia'' (not specific enough). 
    More specifically, if the culture of an assertion contains any of the following words, it will be rejected: \texttt{other}, \texttt{general}, \texttt{1}, \texttt{2}, \texttt{(}, \texttt{)}, \texttt{and}, \texttt{,}, \texttt{some}, \texttt{unknown}, \texttt{parts of}, \texttt{few}, \texttt{/}, \texttt{non-}, \texttt{many}, \texttt{outside}, \texttt{part of}, \texttt{various}, \texttt{elsewhere}, \texttt{rest of}, \texttt{certain}.
\end{itemize}
After filtering, we obtain 507,780 assertions (i.e., 87\% of the original set), 14,298 cultural groups, and 32,126 concepts.

\subsection{Clustering}
For both concept, culture and assertion clustering, we use the HAC algorithm combined with the SentenceBert embedding model \textit{all-MiniLM-L6-v2}\footnote{Model description: \url{https://sbert.net/}}.
We use point-wise Euclidean distance of normalized embeddings, the Ward's linkage~\cite{Ward1963HierarchicalGT}, and the distance threshold set to 1.5, which are adopted from the Candle method~\cite{candle}.

We obtain 4,571 concept clusters and 1,610 culture clusters, with the largest candidate set for assertion clustering containing 463 assertions.
Finally, we obtain 167,396 assertion clusters, where
the largest cluster contains 141 assertions.
Table~\ref{tab:top-10} lists the top cultures and concepts in our dataset.

\begin{table}
\centering
\small
\begin{tabular}{lc|lc}
    \toprule
    \textbf{Concept} & \textbf{Count} & \textbf{Culture} & \textbf{Count} \\
    \midrule
    family & 372 & \multicolumn{2}{c}{\textbf{Countries}}  \\
    tea & 352 & United States & 9097 \\
    hospitality & 349 & Japan & 4597 \\
    personal space & 338 & India & 3670 \\
    marriage & 285 & \multicolumn{2}{c}{\textbf{Regions}}  \\
    family structure & 234 & Western countries & 5788 \\
    education & 228 & Eastern countries & 1849 \\
    gender roles & 200 & Middle Eastern countries & 1059 \\
    public transportation & 193 & \multicolumn{2}{c}{\textbf{Religions}}  \\
    music & 183 & Christianity & 142 \\
    traditional clothing & 180 & Islamic countries & 121 \\
    time & 174 & Hinduism & 106 \\
    food & 149 & \multicolumn{2}{c}{\textbf{Ethnic groups}}  \\
    spicy food & 148 & Amazonian tribes & 143 \\
    communication & 147 & Inuit communities & 101 \\
    fishing & 132 & Maori culture & 73 \\
    \bottomrule
\end{tabular}
\caption{Top concepts and cultures in \ourmethod{}.}
\label{tab:top-10}
\end{table}

\subsection{Costs}
We ran our workflow with the \emph{gpt-3.5-turbo-1106} model in January 2024.
The assertion generation steps costed \$71.
On average, each thousand generated assertions costed 12 cents.
The cluster representative generation costed \$41.
On average, each thousand cluster representative generations costed 25 cents.

The API to the GPT-3.5 model we used in this implementation has limits at 1M input tokens per minute, and 10K requests per minute. By making concurrent requests, our workflow could be executed in less than 5 hours, including the clustering steps which took less than 30 minutes.

\section{Quality of \ourmethod{} Assertions}
\label{sec:intrinsic-evaluation}

We evaluate the assertion quality of \ourmethod{} by comparison to several resources of similar kind via human annotation.

\subsection{Compared Resources}
Our baseline resources are Quasimodo~\cite{quasimodo}, StereoKG~\cite{stereokg} and Candle~\cite{candle}.
StereoKG and Candle are specialized in culture-specific knowledge.
Quasimodo is a general commonsense knowledge base, but for this comparison, we only consider Quasimodo assertions that contain geo-locations or religions.

StereoKG and Candle store assertions in the form of natural-language sentences, while Quasimodo assertions originally come in triple format (subject-predicate-object).
Therefore, for comparability, we verbalized the Quasimodo assertions using GPT-3.5.

For our resource, in addition to the full assertion set (\ourmethodfull), we also evaluate \ourmethodtop, which contains the most frequent assertions up to the same size as Candle (i.e., 60K assertions).

\paragraph{Other resources}
There is no publicly available resource from the work of \citet{Anacleto2006}.
Similarly, the GD-COMET model \cite{bhatia-shwartz-2023-gd} has not been published.
The resource by \citet{acharya2020towards} only consists of 225 assertions for two cultures (USA and India), which has very limited size and scope compared to other resources, hence it is not included in our evaluation.
We acknowledge that contemporaneous works such as NormsKB \cite{fung-etal-2023-normsage} and CultureAtlas \cite{fung2024massively} are comparable to our method, and will include them in future studies.

\subsection{Metrics}
CCSK assertions should satisfy a set of conditions, that are sometimes in conflict. We employ three dimensions to evaluate their intrinsic quality, influenced by \citet{bhatia-shwartz-2023-gd}.
\begin{enumerate}
    \item \textbf{Cultural Relevance (CR)} measures the plausibility of the assertion, i.e., whether the statement actually represents cultural knowledge.
    \item \textbf{Stereotype Avoidance (SA)} assesses if the assertion avoids reinforcing cultural stereotypes and presenting offensive materials.
    \item \textbf{Linguistic Accuracy (LA)} checks if the assertion is grammatical, meaningful and understandable as a standalone sentence.
\end{enumerate}


\subsection{Setup}
Our evaluation is conducted on the Amazon Mechanical Turk (MTurk) platform.
Specifically, we ask human annotators on MTurk to evaluate each assertion along the three dimensions (CR, SA, and LA), each on a 3-point Likert scale: \emph{negation} (0), \emph{ambiguity} (1) and \emph{affirmation} (2).

For each resource, we randomly sampled 300 assertions for evaluation,
each assessed by 3 different workers.
Each MTurk HIT contained 5 assertions and was compensated \$0.30.
We estimated each HIT would take 1.5 minutes to complete, hence an hourly wage of \$12.
The annotators were required to be Master Workers with a lifetime approval rate greater than 99\% (this figure is provided by the platform).
There were totally 33 different workers who participated in this task.
We observed the following Krippendorff’s Alpha values for each of the metrics:
Cultural Relevance: 0.33,
Stereotype Avoidance: 0.29,
Linguistic Accuracy: 0.39.

As part of the MTurk tasks, we also asked annotators to specify how familiar they are with the cultures mentioned in the assertions. 41.8\% of the annotators reported being ``very familiar'' with the cultures presented in the tasks, 37.4\% being ``somewhat familiar'', and 20.8\% being ``not familiar''. In the latter case, our guidelines advised the annotators to search and inspect additional sources on the Internet.

\subsection{Results}

\begin{table}
\centering

\begin{tabular}{lccccc}
\toprule
\multirow{3}{*}{\textbf{Resource}} & \multicolumn{3}{c}{\textbf{Quality ({[}0..2{]}})} & \multicolumn{2}{c}{\textbf{Size}} \\
\cmidrule(lr){2-4} \cmidrule(lr){5-6}
& \textbf{CR} & \textbf{SA} & \textbf{LA} & \textbf{\#C} & \textbf{\#A} \\
\midrule
\textbf{StereoKG}~\cite{stereokg} & 0.79 & 0.85 & 1.11 & 10 & 4K\\
\textbf{Quasimodo}~\cite{quasimodo} & 0.85 & 1.22 & 1.12 & 0.4K & 131K \\
\textbf{Candle}~\cite{candle} & 1.42 & 1.54 & 1.67 & 0.4K & 60K\\
\rowcolor[gray]{.9}
\textbf{\ourmethodfull} & 1.53* & 1.61* & 1.79* & \textbf{11.1K} & \textbf{167K}\\
\rowcolor[gray]{.9}
\textbf{\ourmethodtop} & \textbf{1.59*} & \textbf{1.65*} & \textbf{1.82*} & 8.1K & 60K\\
\bottomrule
\end{tabular}

\caption{Results of \emph{intrinsic evaluation} via human annotations (metrics: \textbf{CR} - Cultural Relevance, \textbf{SA} - Stereotype Avoidance, \textbf{LA} - Linguistic Accuracy), along with the coverage of the resources (\textbf{\#C} - number of cultures, \textbf{\#A} - number of assertions). The (*) marker indicates results with statistically significant gains over Candle ($p < 0.05$ in the Student's t-test). }
\label{tab:intrinsic-eval}
\end{table}

We report average scores given by the annotators in Table~\ref{tab:intrinsic-eval}.
It can be seen that \ourmethod{} assertions have the highest quality compared to prior resources.
Notably, despite being nearly 3 times larger than the largest baseline, \ourmethodfull{} has better quality than all baselines, on all metrics.

Moreover, our top-ranked assertion subset, \ourmethodtop{}, has the same size as Candle, but it outperforms this resource on all 3 metrics by a large margin.
This affirms that the frequency signals we obtain in our clustering step are helpful for pulling out the most significant assertions.

\section{CCSK for Intercultural Dialogues}
\label{sec:extrinsic-evaluation}


Explicit knowledge has been used to improve LLM performance in downstream tasks such as social dialogue generation~\cite{kim-etal-2023-soda} and intra-cultural dialogue synthesis~\cite{li-etal-2023-normdial}.
However, other important scenarios, which involve \emph{intercultural} interactions, remain under-explored.
To address this gap, 
we introduce two new tasks concerning intercultural conversations.
We will show that explicit CCSK assertions from resources like \ourmethod{} can improve LLM responses in dialogue applications requiring cross-cultural knowledge.

\subsection{Next Utterance Generation}
\label{subsec:next-utterance-gen}

The input of this task includes: (1) a narrative that involves interactions between two people of different cultures, and (2) their ongoing dialogue.
The task is to generate the utterance of the next speaker.
To excel in this task, LLMs need to leverage cultural knowledge such that the generated responses are natural, consistent, culturally sensitive and specific.
%
We provide an example of input and output for this task in Table~\ref{tab:extrinsic-task-compact-example}.

\begin{table}
    \centering
    \small
    \begin{tabular}{p{.96\columnwidth}}
    \toprule
    \multicolumn{1}{c}{\textbf{Input}} \\
    \midrule
    {\cellcolor[gray]{.9}\textbf{Narrative}} \\
    John, an American, is visiting his friend Kenji, who lives in Tokyo. They are paying their bill for dinner at a restaurant. \\
    \midrule
    {\cellcolor[gray]{.9}\textbf{Ongoing dialogue}} \\
    \textbf{John:} That's a great meal, Kenji. I really liked the sushi. \\
    \textbf{Kenji:} My pleasure, John. I'm glad you enjoyed it. \\
    \textbf{John:} Let me see the bill. It is 8,000 yen. I'm gonna leave 10,000 yen. \\
    \bottomrule \toprule
    \multicolumn{1}{c}{\textbf{Output}} \\
    \midrule
    {\cellcolor[gray]{.9}\textbf{Method 1: Vanilla GPT-3.5}} \\
    \textbf{Kenji:} Thank you, John. You're too kind. Next time, dinner is on me. It's a very generous tip too. \\
    \midrule
    {\cellcolor[gray]{.9}\textbf{Method 2: GPT-3.5 with explicit CCSK}} \\
    {\cellcolor[gray]{.9}\textbf{Injected CCSK:} \textit{Tipping is not a common practice in Japan and can be considered rude or impolite.}} \\
    \textbf{Kenji:} Oh, no, John. You don't need to leave a tip here in Japan. Just 8,000 yen is fine. Thank you for offering though. \\
    \bottomrule
    \end{tabular}
    \caption{Examples for the \textit{next utterance generation} task.
    }
    \label{tab:extrinsic-task-compact-example}
\end{table}

\begin{table*}
    \centering
    \footnotesize
    \begin{tabular}{l|l}
        \toprule
        \textbf{Input} & \begin{tabular}[c]{p{0.88\textwidth}}You are a narrative generator. Your task is to generate short narratives of less than 5 sentences around a cultural concept that involves two people from two different cultures. \textbf{The narrative should lead to an intercultural interaction where cultural differences play a significant role.} You must not include the cultural differences, or cultural knowledge, or the resolution, or the consequences of the situation in the narrative. \\
\\
Some examples: \\
- Anna, an American, is visiting a village in Vietnam where Minh is a local. Anna asks Minh where she can get food for her dog. \\
- Erling from Norway is visiting Seoul. He and his new friend, Heungmin, are picking foods for their dinner at a traditional restaurant. \\
- Liz and Qiang are two friends, who are currently in England. Qiang is from China who is visiting the country. Liz is a local. They are preparing tea together. \\
\\
Please write 3 more narratives: \end{tabular}\\
        \midrule
        \textbf{Output} & \begin{tabular}[c]{p{0.88\textwidth}}- Maria, a Mexican, is visiting Japan and is shopping in a local market. She meets Yuki, a Japanese woman, and asks for help in choosing a traditional Japanese outfit for a festival.\\
- Pablo, a Spaniard, is traveling in India and meets Rajesh, a local, at a temple. They both want to participate in a religious ceremony, and Pablo asks Rajesh for guidance on the proper etiquette.\\
- Fatima, a Saudi Arabian, is studying in France and meets Pierre, a French student, at a party. They both want to dance, and Fatima asks Pierre to teach her a traditional French dance. \end{tabular} \\
        \bottomrule
    \end{tabular}
    \caption{An example of prompts used to generate intercultural narratives for the two extrinsic tasks.}
    \label{tab:narrative-gen}
\end{table*}

\begin{table}
    \centering
    \small
    \begin{tabular}{p{.96\columnwidth}}
    \toprule
    {\cellcolor[gray]{.9}\textbf{Example 1}} \\
    \textbf{Narrative:} Carlos from Argentina is visiting Korea. He greets his new Korean friend, Jihoon, by giving him a friendly pat on the back. \\
    \textbf{Query:} X from Argentina is visiting Korea. He greets his new Korean friend, Y, by giving him a friendly pat on the back. \\
    \textbf{Retrieval Results:} \\
    • In South Korea, beckoning with an open hand or palm facing downwards is considered polite. \textit{(Similarity: 0.5216)} \\
    • In South Korean culture, it is common to gently pat someone on the shoulder or back as a sign of encouragement or reassurance. \textit{(Similarity: 0.5201)} \\
    \toprule
    \toprule
    {\cellcolor[gray]{.9}\textbf{Example 2}} \\
    \textbf{Narrative:} Maria, a woman from Spain, is visiting a Bedouin tribe in Jordan upon an invitation from her new friend, Ahmed. They are preparing to have dinner under the star-lit desert sky. \\
    \textbf{Query:} X, a woman from Spain, is visiting a Bedouin tribe in Jordan upon an invitation from her new friend, Y. They are preparing to have dinner under the star-lit desert sky. \\
    \textbf{Retrieval Results:} \\
    • Bedouins, a Middle Eastern culture, are known for their nomadic lifestyle, hospitality, and expertise in desert survival. \textit{(Similarity: 0.5217)} \\
    • Desert cultures highly value hospitality and express it through offering food, drinks, and shelter to guests. \textit{(Similarity: 0.5207)} \\
    \bottomrule
    \end{tabular}
    \caption{Examples of retrieving \ourmethod{}'s CCSK assertions relevant to given narratives.}
    \label{tab:ccsk-retrieval-example}
\end{table}

\subsubsection{Input Construction}
First, using GPT-3.5 and a small set of few-shot examples, we generate short narratives involving two people of different cultures, e.g., \textit{``Liz from England is visiting her friend Qiang in China. They are preparing tea together.''}
(see in Table~\ref{tab:narrative-gen} for a prompt sample).

For each narrative, we ask the LLM to generate a possible dialogue between the two participants in that situation.
The first three turns of each generated dialogue are used to construct test samples.

\subsubsection{Methods}
We compare two methods, without and with explicit CCSK assertions in the prompts.
\begin{enumerate}
    \item  \emph{Method 1 (Vanilla LLM)}
    prompts an LLM with the task description, the input narrative, and the dialogue history.
    \item \emph{Method 2 (LLM with explicit CCSK)}
    alters the prompts of \emph{Method 1} by augmenting relevant CCSK.
\end{enumerate}

\paragraph{CCSK Retrieval}
For \emph{Method 2}, we use a simple dense embeddings retrieval approach.
First, we compute the embeddings of all assertions in our \ourmethod{} dataset using the SentenceBert model \emph{all-MiniLM-L6-v2}.
Then, given a narrative, we replace the names of the people with ``X'' and ``Y'' in order to reduce the distractions for the embedding model.
For each given narrative, we retrieve the top-2 most similar CCSK assertions from \ourmethod{} based on cosine similarity of the embeddings.
We also restrict the similarity to be better than 0.5.
We only consider narratives that have relevant CCSK assertions retrieved from \ourmethod{} in our evaluation.
Examples of CCSK retrieval can be seen in Table~\ref{tab:ccsk-retrieval-example}.

\begin{figure*}
    \centering
    \footnotesize
    \begin{subfigure}{\textwidth}
        \centering
        \caption*{
        \tiny
            \ref{bar:vanilla-llm} Vanilla LLM 
            \enspace
            \ref{bar:llm-ccsk} LLM w/ explicit CCSK
        }
    \end{subfigure}
    \hfill
    \begin{subfigure}{.33\textwidth}
        \begin{tikzpicture}
        \begin{axis}[
            ybar,
            enlargelimits=0.15,
            symbolic x coords={Naturalness, Consistency, Specificity, Cultural Sensitivity*, Overall Quality},
            xtick=data,
            x tick label style={rotate=30,anchor=east},
            bar width=0.15cm,
            height=3.0cm,
            width=\columnwidth,
            ymin=20,
            ymax=50,
            ytick={20,35,50},
            yticklabels={20\%,35\%,50\%},
            ]
        \addplot[black,pattern=north east lines]
            coordinates {(Naturalness,36.7) (Consistency,30.3) 
                (Specificity,38.0) (Cultural Sensitivity*,28.7) (Overall Quality,40.0)}; \label{bar:vanilla-llm}
        \addplot[teal,fill=teal]
            coordinates {(Naturalness,37.3) (Consistency,35.0) 
                (Specificity,47.0) (Cultural Sensitivity*,40.3) (Overall Quality,50.3)}; \label{bar:llm-ccsk}
        \end{axis}
        \end{tikzpicture}
        \caption{GPT}%
    \label{subfig:eval-next-utterance-gpt}
    \end{subfigure}
    \hfill
    \begin{subfigure}{.33\textwidth}
        \begin{tikzpicture}
        \begin{axis}[
            ybar,
            enlargelimits=0.15,
            symbolic x coords={Naturalness, Consistency, Specificity, Cultural Sensitivity*, Overall Quality},
            xtick=data,
            x tick label style={rotate=30,anchor=east},
            bar width=0.15cm,
            height=3.0cm,
            width=\columnwidth,
            ymin=20,
            ymax=50,
            ytick={20,35,50},
            yticklabels={20\%,35\%,50\%},
            ]
        \addplot[black,pattern=north east lines]
            coordinates {(Naturalness,30.7) (Consistency,33.3) 
                (Specificity,37.0) (Cultural Sensitivity*,27.3) (Overall Quality,40.0)};
        \addplot[teal,fill=teal]
            coordinates {(Naturalness,32.3) (Consistency,32.0) 
                (Specificity,41.3) (Cultural Sensitivity*,39.7) (Overall Quality,45.3)};
        \end{axis}
        \end{tikzpicture}
        \caption{Mistral}%
    \label{subfig:eval-next-utterance-mistral}
    \end{subfigure}
    \hfill
    \begin{subfigure}{.33\textwidth}
        \begin{tikzpicture}
        \begin{axis}[
            ybar,
            enlargelimits=0.15,
            symbolic x coords={Naturalness, Consistency, Specificity*, Cultural Sensitivity, Overall Quality*},
            xtick=data,
            x tick label style={rotate=30,anchor=east},
            bar width=0.15cm,
            height=3.0cm,
            width=\columnwidth,
            ymin=20,
            ymax=50,
            ytick={20,35,50},
            yticklabels={20\%,35\%,50\%},
            ]
        \addplot[black,pattern=north east lines]
            coordinates {(Naturalness,33.7) (Consistency,22.3) 
                (Specificity*,22.7) (Cultural Sensitivity,18.0) (Overall Quality*,34.0)};
        \addplot[teal,fill=teal]
            coordinates {(Naturalness,40.7) (Consistency,29.7) 
                (Specificity*,48.3) (Cultural Sensitivity,24.7) (Overall Quality*,50.3)};
        \end{axis}
        \end{tikzpicture}
        \caption{Bard}%
    \label{subfig:eval-next-utterance-bard}
    \end{subfigure}
    \caption{
    Results of head-to-head comparison for the \emph{next utterance generation} task 
    via human evaluation. 
    The y-axis represents the percentage of samples preferred by the evaluators (the ``Tie'' cases are omitted from the plots).
    The (*) marker indicates results with statistically significant differences
    ($p < 0.05$ in the Wilcoxon T-test).
    }
    \Description{Bar charts showing the superiority of LLMs + explicit CCSK against Vanilla LLM.}
    \label{fig:next-utterance-evaluation-results}
\end{figure*}
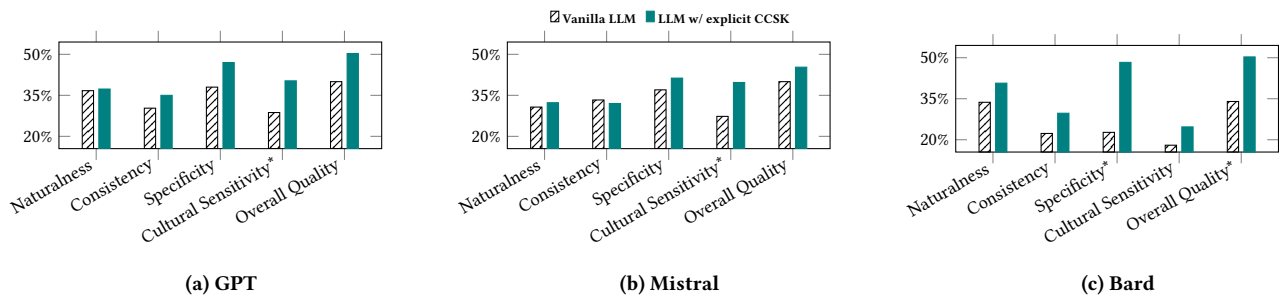

\subsubsection{Setup}
We experiment with three different LLMs from some of the most popular model families: (i) GPT-3.5\footnote{Model name: \textit{gpt-3.5-turbo-1106}}, also known as ChatGPT, by OpenAI,
(ii) the strongest model of the Mistral family by MistralAI\footnote{Model name: \textit{mistral-medium} (\url{https://docs.mistral.ai/platform/endpoints/})}, and (iii) Google's largest globally accessible language model as of January 2024, Bard\footnote{Accessed via \url{https://bard.google.com/}}.
For each LLM, we randomly draw 100 test samples and use the two methods to generate 100 pairs of responses.
We employ human annotators on MTurk to evaluate the quality of the generated 
utterances.
Specifically,
we show side-by-side the outputs of the two methods given the same input narrative, and ask the evaluators to choose the better one based on the following dimensions:
\begin{enumerate}
    \item \textbf{Naturalness:} The response does not sound awkward or unnatural.
    \item \textbf{Consistency:} The response does not contradict the narrative or previous utterances.
    \item \textbf{Specificity:} The response contains specific details rather than vague/generic information.
    \item \textbf{Cultural Sensitivity:} The response shows respect and understanding of the cultures.
    \item \textbf{Overall Quality:} The overall satisfaction with the response.
\end{enumerate}
Dimensions 1-3 and 5 are established criteria for evaluating dialogues~\cite{mehri2022report, kim-etal-2023-soda}, dimension 4 is a dimension we introduce in this evaluation, as per our focus on cross-cultural knowledge.


Each sample is evaluated by three different MTurk workers.
To alleviate positional bias, the orders of the outputs are randomly shuffled before being revealed to the annotators.
The instruction and layout of our annotation page used for MTurk evaluations are influenced by the work of \citet{kim-etal-2023-soda}.
Instead of using ``Definitely A/B'' and ``Slightly A/B'', we narrow the option set down to ``A is Better'', ``B is Better'', and introduce a third option which is ``Tie''.
For these HITs, we require the workers to be Master Workers who have a lifetime approval rate of more than 99\%.

There were 41 different MTurk workers who participated in this evaluation.
Each HIT was compensated \$0.20.
We estimated each HIT would take 1 minute to complete, hence an hourly wage of \$12.
In these MTurk tasks,
40.9\% of the annotators reported being ``very familiar'' with the presented cultures,
48.2\% being ``somewhat familiar'',
and
10.9\% being ``not familiar''.

\subsubsection{Results}

The human evaluation results are demonstrated in Figure~\ref{fig:next-utterance-evaluation-results}.
It can be seen that explicit CCSK assertions help all three LLMs to perform significantly better in this task, as the \emph{overall quality} of the generated utterances are preferred by the human annotators compared to the outputs of the vanilla LLMs.
This is attributed by the fact that the CCSK-enhanced utterances are more \emph{specific} and \emph{culturally sensitive} (for all three LLMs), and even more \emph{natural} and \emph{consistent} (for Bard).
This shows that even though LLMs inherently possess cultural knowledge, they often fail to incorporate such knowledge into the generated utterances, and that explicit CCSK injected in the prompts can mitigate this problem.
We provide some evaluated samples in Table~\ref{tab:mturk-utterance-examples}.

\subsubsection{\ourmethod{} vs. Candle}
Additionally, we compared using \ourmethod{} vs.\ Candle~\cite{candle} assertions for augmenting prompts to GPT-3.5.
We sample 100 test narratives and use the same retrieval method for both resources in this comparison.
The results are presented in Table~\ref{tab:next-utterance-candle-vs-mango}.
It can be seen that human annotators preferred utterances generated with \ourmethod{} assertions in terms of naturalness, consistency and specificity.
However, we note that the size of the test set is very small, and the difference in the results is not statistically significant.
A larger-scale evaluation with a broader scope is needed to evaluate CCSK resources more comprehensively.

\begin{table}
    \centering
    \begin{tabular}{lcc}
    \toprule
    \textbf{Metric} & \textbf{Candle}~\cite{candle} & \textbf{\ourmethod{}} \\
    \midrule
    \textbf{Naturalness} & 30.7\% & \textbf{32.7\%} \\
    \textbf{Consistency} & 28.0\% & \textbf{30.7\%} \\
    \textbf{Specificity} & 37.0\% & \textbf{39.7\%} \\
    \textbf{Cultural Sensitivity} & 34.3\% & 34.3\% \\
    \textbf{Overall Quality} & 46.0\% & 46.0\% \\
    \bottomrule
    \end{tabular}
    \caption{Results of head-to-head human evaluation for the \emph{next utterance generation} task using GPT-3.5, with Candle vs. \ourmethod{} assertions injected in the prompts. The numbers present the percentage of samples preferred by evaluators.}
    \label{tab:next-utterance-candle-vs-mango}
\end{table}

\subsection{Full Dialogue Generation}
\label{subsec:full-dialog-gen}

In this task, instead of generating a single utterance, we ask LLMs to generate a full dialogue between the two people in a given narrative.
We anticipate that this task is more challenging both for LLMs to solve and for human annotators to evaluate.

We randomly sample 100 test narratives and ask GPT-3.5 to generate dialogues.
We use the same prompting methods and evaluation setup as in the previous task.
The evaluation results can be found in Table~\ref{tab:full-dialog-evaluation-results}.
It can be seen that injecting explicit CCSK in the prompts does not diminish the natural flow and consistency of the generated dialogues, on the other hand, it provides more specificity and cultural sensitivity into the intercultural interactions.
As a result, the annotators prefer the CCSK-enhanced dialogues on overall quality, similar to the results of the next utterance generation task.

There were 29 different MTurk participants in this evaluation.
Each HIT was compensated \$0.30.
In these tasks,
33.7\% of the annotators reported being ``very familiar'' with the presented cultures,
54.0\% being ``somewhat familiar'',
and
12.3\% being ``not familiar''.

\begin{table}
    \centering
    \begin{tabular}{lcc}
    \toprule
    \textbf{Metric} & \textbf{No CCSK} & \textbf{With CCSK} \\
    \midrule
    \textbf{Naturalness} & \textbf{39.3\%} & 38.7\% \\
    \textbf{Consistency} & 21.0\% & \textbf{22.7\%} \\
    \textbf{Specificity} & 36.7\% & \textbf{43.0\%} \\
    \textbf{Cultural Sensitivity} & 22.7\% & \textbf{28.3\%} \\
    \textbf{Overall Quality} & 40.7\% & \textbf{44.7\%} \\
    \bottomrule
    \end{tabular}
    \caption{Results of head-to-head human evaluation for the \emph{full dialogue generation} task with GPT-3.5. The numbers present the percentage of samples preferred by evaluators.}
    \label{tab:full-dialog-evaluation-results}
\end{table}

\begin{table*}
    \centering
    \footnotesize
    \begin{tabular}{p{.14\textwidth}|p{.14\textwidth}|p{.14\textwidth}|p{.14\textwidth}|p{.14\textwidth}|p{.14\textwidth}}
    \toprule
         \multicolumn{6}{c}{\textit{\textbf{Sample 1}}}  \\
    \midrule
        \multicolumn{6}{G}{\textbf{Narrative}} \\
        \multicolumn{6}{c}{\begin{tabular}[c]{p{0.87\textwidth}}Ali from Tunisia is visiting Tokyo for the first time. He and his newly made Japanese friend, Hiroshi, are deciding how to celebrate Ali's birthday in one of Tokyo's iconic districts.\end{tabular}} \\
    \midrule
        \multicolumn{6}{G}{\textbf{Dialogue}} \\
        \multicolumn{6}{c}{\begin{tabular}[c]{p{0.87\textwidth}}Ali: Hiroshi, I'm so excited to celebrate my birthday in Tokyo! What do you think we should do?\\
Hiroshi: Happy birthday, Ali! I think we should start with a traditional Japanese meal at a local izakaya.\\
Ali: That sounds great! I've been wanting to try authentic Japanese cuisine. And then maybe we could check out the vibrant nightlife in Shibuya?\end{tabular}} \\
    \midrule
        \multicolumn{3}{G|}{\textbf{Vanilla GPT}} & \multicolumn{3}{G}{\textbf{\ourgpt{}}} \\
    \midrule
        \multicolumn{3}{l|}{\begin{tabular}[c]{p{0.415\textwidth}}-\end{tabular}} & \multicolumn{3}{l}{\begin{tabular}[c]{p{0.415\textwidth}}\textbf{Injected CCSK:}\\
        • In Japan, birthdays are celebrated with symbolic rituals, special meals, and a focus on traditional ceremonies and gift-giving.\end{tabular}} \\
    \midrule
         \multicolumn{3}{l|}{\begin{tabular}[c]{p{0.415\textwidth}}Hiroshi: Absolutely, Ali! Shibuya is the perfect place to experience Tokyo's bustling nightlife. We could start with a few drinks at a traditional Japanese izakaya and then explore the neon-lit streets and vibrant energy of Shibuya. It's going to be an unforgettable birthday celebration!\end{tabular}} & \multicolumn{3}{l}{\begin{tabular}[c]{p{0.415\textwidth}}Hiroshi: That sounds like a fantastic plan, Ali! But I also think it would be wonderful to incorporate some Japanese birthday traditions into our celebration. How about we visit a shrine to make a prayer for good luck and then participate in the Omikuji fortune-telling ritual? It's a special way to mark your birthday and immerse yourself in Japanese culture. And of course, we can still enjoy the bustling nightlife in Shibuya afterwards.\end{tabular}} \\
     \midrule
        \rowcolor[gray]{0.9}
        \textbf{Criterion} & \textbf{Natural Flow} & \textbf{Consistency} & \textbf{Specificity} & \textbf{Cultural Sensitivity} & \textbf{Overall Quality} \\
        \midrule
        \textbf{Annotator 1} & \ourgpt{} & Tie & \ourgpt{} & \ourgpt{} & \ourgpt{} \\
        \midrule
        \textbf{Annotator 2} & Tie & Tie & \ourgpt{} & Tie & \ourgpt{} \\
        \midrule
        \textbf{Annotator 3} & Vanilla GPT & \ourgpt{} & \ourgpt{} & \ourgpt{} & \ourgpt{} \\
        
    \bottomrule
    \toprule
    
         \multicolumn{6}{c}{\textit{\textbf{Sample 2}}}  \\
    \midrule
        \multicolumn{6}{G}{\textbf{Narrative}} \\
        \multicolumn{6}{c}{\begin{tabular}[c]{p{0.87\textwidth}}Fiona, a woman from Scotland, is in Nairobi visiting her online friend Makena. Makena is taking her to a traditional Kenyan hair salon for the first time.\end{tabular}} \\
    \midrule
        \multicolumn{6}{G}{\textbf{Dialogue}} \\
        \multicolumn{6}{c}{\begin{tabular}[c]{p{0.87\textwidth}}Makena: Welcome to the salon, Fiona! Are you excited to try out some Kenyan hair styles?\\
Fiona: Oh, I am a bit nervous, but also curious to see what kind of styles are popular here. It's quite different from what I'm used to in Scotland.\\
Makena: Don't worry, you're going to love it! Kenyan hair stylists are really talented and they have some unique techniques.\end{tabular}} \\
    \midrule
        \multicolumn{3}{G|}{\textbf{Vanilla GPT}} & \multicolumn{3}{G}{\textbf{\ourgpt{}}} \\
    \midrule
        \multicolumn{3}{l|}{\begin{tabular}[c]{p{0.415\textwidth}}-\end{tabular}} & \multicolumn{3}{l}{\begin{tabular}[c]{p{0.415\textwidth}}\textbf{Injected CCSK:}\\
        • Hairdressing education in Kenya focuses on intricate braiding and natural hair care techniques for diverse African hair types and textures. \\
        • In Kenya, elaborate and intricate braided hairstyles are a significant aspect of cultural identity.\end{tabular}} \\
    \midrule
         \multicolumn{3}{l|}{\begin{tabular}[c]{p{0.415\textwidth}}Fiona: That's good to hear! I can't wait to see what they can do. I've always admired African hairstyles from afar, so I'm really looking forward to experiencing it firsthand.\end{tabular}} & \multicolumn{3}{l}{\begin{tabular}[c]{p{0.415\textwidth}}Fiona: That sounds amazing! I've always admired the intricate braided hairstyles I've seen here. I've never had the chance to experience it myself, so I'm really looking forward to it.\end{tabular}} \\
     \midrule
        \rowcolor[gray]{0.9}
        \textbf{Criterion} & \textbf{Natural Flow} & \textbf{Consistency} & \textbf{Specificity} & \textbf{Cultural Sensitivity} & \textbf{Overall Quality} \\
        \midrule
        \textbf{Annotator 1} & Tie & \ourgpt{} & \ourgpt{} & Tie & \ourgpt{} \\
        \midrule
        \textbf{Annotator 2} & \ourgpt{} & \ourgpt{} & \ourgpt{} & \ourgpt{} & \ourgpt{} \\
        \midrule
        \textbf{Annotator 3} & \ourgpt{} & \ourgpt{} & \ourgpt{} & \ourgpt{} & \ourgpt{} \\
    \bottomrule
    
    \end{tabular}
    \caption{Examples of responses generated by GPT-3.5 without and with augmented CCSK, and the annotations of MTurk workers in our evaluation.}
    \label{tab:mturk-utterance-examples}
\end{table*}

\section{Discussion} \label{sec:discussion}
We discuss the shortcomings of this work and future opportunities along this line of research.

\paragraph{Knowledge Base Construction Using LLMs}
Using LLMs
to generate facts for knowledge bases has become a major research direction.
Our work follows this idea with a focus on cultural knowledge.
Although our proposed method for CCSK distillation outperforms text extraction by a large margin in terms of quality and size, there are three inherent limitations when using GPT-3.5:
\begin{enumerate}
    \item As GPT-3.5 is provided via a commercial API only, it cannot be guaranteed that our pipeline is long-term reproducible.
    \item Since the training process of GPT-3.5 is undocumented, we do not know what corpus it has been trained on, and in particular, whether there might be systematic gaps in its training corpus (know what you don't know).
    \item Not specific to GPT-3.5, the usage of LLMs for CCSK distillation means losing the ability to trace assertions to their source, compared with text extraction. 
\end{enumerate}

\paragraph{Stereotypes}
Adapting conversational AI to cultural contexts is important for opening systems to diverse communities. At the same time, CCSK represents dual-use technology, as it can also be used to reinforce stereotypes, and discriminate.

As generative AI is getting increasingly popular, stereotype detection has become a notable research area.
There have been several efforts to develop datasets for stereotype classification (e.g., \cite{dinan-etal-2020-multi,felkner-etal-2023-winoqueer,nangia-etal-2020-crows,nadeem-etal-2021-stereoset}).
Such datasets can be used to train stereotype detection models which can be helpful for downstream applications to select which assertions should be incorporated into their systems.

Another possibility is to trace the source of generated assertions, for example, via web search.
Then, the source websites can be ranked by their trustworthiness, giving signals for stereotype detection, besides the frequency signals given by the consolidation step in our workflow.

Nevertheless, our data represents the outputs of a \emph{research prototype}. We recommend against using our data in production systems without a careful evaluation of benefits (of having a system that is more culturally adept) and associated challenges and risks (of stereotype perpetuation etc.).

\paragraph{Human Evaluation in Cross-Cultural Tasks}
It has been reported that the majority of MTurk workers are from only two cultures (USA and India) \cite{10.1145/1753846.1753873}, which makes it challenging to recruit truly diverse annotators.
In our evaluation, 
although the majority of the workers reported being somewhat or very familiar with the cultures, it is not a guarantee for the quality of the annotations.
Conducting evaluations with more culturally diverse annotators would further reinforce our results, but it can be challenging due to the cost and time constraints.
We leave this for future work.

\section{Conclusion}
We presented \ourmethod{}, a methodology to distill and consolidate CCSK from LLMs.
We executed the \ourmethod{} workflow using GPT-3.5, obtaining a collection of 167K assertions covering 11K cultures and 30K concepts, which surpasses existing resources in quality and coverage by a large margin
according to human annotators.
In the extrinsic evaluation, we showed that by augmenting explicit assertions from \ourmethod{} into prompts, LLMs can perform better in intercultural dialogue generation tasks as their responses were judged more specific, more culturally sensitive, and better overall quality by human evaluators.
Our code and data are available at \website{}.






\bibliographystyle{ACM-Reference-Format}
\balance
\bibliography{refs}

\end{document}